\documentclass[runningheads]{llncs}

\usepackage{eccv}

\usepackage{eccvabbrv}

\usepackage{graphicx}
\usepackage{wrapfig}
\usepackage{booktabs}
\usepackage{subcaption}

\usepackage[accsupp]{axessibility}  

\usepackage{hyperref}

\usepackage[dvipsnames]{xcolor}
\usepackage{pgfplots}
\usepgfplotslibrary{groupplots}
\usepackage{tabularray}
\usepackage{adjustbox}

\usepackage{orcidlink}

\begin{document}

\title{I-MedSAM: Implicit Medical Image Segmentation with Segment Anything} 

\titlerunning{I-MedSAM: Implicit Medical Image Segmentation with Segment Anything}

\author{Xiaobao Wei\inst{1,2,3,*}\orcidlink{0000-0003-4230-1162} \and
Jiajun Cao\inst{1,4,*} \and
Yizhu Jin\inst{1,5} \and \\
Ming Lu\inst{1} \and
Guangyu Wang\inst{6} \and
Shanghang Zhang\inst{1,\dagger}
}

\authorrunning{X. Wei et al.}

\newlength{\nameblob}
\settowidth{\nameblob}{ResUNet2}
\newlength{\blob}
\settowidth{\blob}{Mi}
\institute{
$^1$State Key Laboratory of Multimedia Information Processing, School of Computer Science, Peking University $^2$University of Chinese Academy of Sciences \\
$^3$Institute of Software, Chinese Academy of Sciences $^4$Xi'an Jiaotong University \\
$^5$Beihang University $^6$Beijing University of Posts and Telecommunications 
\email{weixiaobao0210@gmail.com}
}

\maketitle

\begin{abstract}
With the development of Deep Neural Networks (DNNs), many efforts have been made to handle medical image segmentation. Traditional methods such as nnUNet train specific segmentation models on the individual datasets. Plenty of recent methods have been proposed to adapt the foundational Segment Anything Model (SAM) to medical image segmentation. However, they still focus on discrete representations to generate pixel-wise predictions, which are spatially inflexible and scale poorly to higher resolution. In contrast, implicit methods learn continuous representations for segmentation, which is crucial for medical image segmentation. In this paper, we propose I-MedSAM, which leverages the benefits of both continuous representations and SAM, to obtain better cross-domain ability and accurate boundary delineation. Since medical image segmentation needs to predict detailed segmentation boundaries, we designed a novel adapter to enhance the SAM features with high-frequency information during Parameter-Efficient Fine-Tuning (PEFT). To convert the SAM features and coordinates into continuous segmentation output, we utilize Implicit Neural Representation (INR) to learn an implicit segmentation decoder. We also propose an uncertainty-guided sampling strategy for efficient learning of INR. Extensive evaluations on 2D medical image segmentation tasks have shown that our proposed method with only 1.6M trainable parameters outperforms existing methods including discrete and implicit methods. The code will be available at: \href{https://github.com/ucwxb/I-MedSAM}{https://github.com/ucwxb/I-MedSAM}.
\end{abstract}

\renewcommand{\thefootnote}{\fnsymbol{footnote}} 
\footnotetext[1]{Equal Contribution.} 
\footnotetext[4]{Corresponding Author.}

\section{Introduction}

\begin{figure}[tb]
  \centering
  \begin{subfigure}{0.46\linewidth}
    \includegraphics[width=\linewidth]{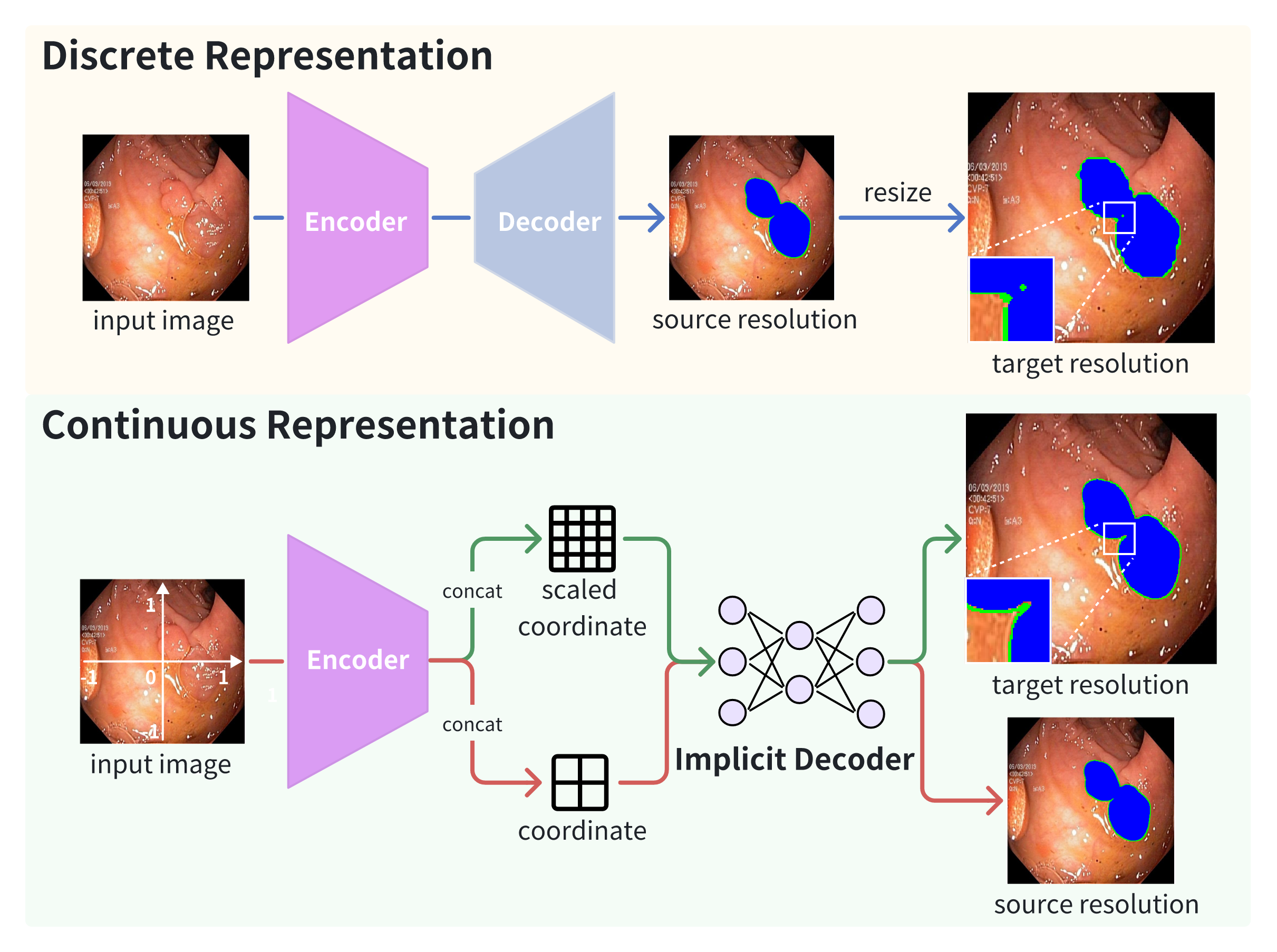}
    \caption{Illustrations of discrete and continuous representation for medical segmentation.}
    \label{fig:short-a}
  \end{subfigure}
  \hfill
  \begin{subfigure}{0.52\linewidth}
    \begin{tikzpicture}[scale=0.55]
    \begin{axis}[
    ybar,ymin=65,
    ytick={65, 70, 75, 80, 85, 90, 100},
    width=2.0\linewidth, 
    enlarge x limits=0.2,
    height=6cm,
    ylabel=$Dice$ (\%),
    legend style={at={(1.0,-0.25), anchor=north}, legend columns=-1},
    symbolic x coords={Source-domain, Cross-domain, Lower-resolution, Higher-resolution},
    y label style={at={(axis description cs:0.15,1.0)}, anchor=south, rotate=270}, 
    xtick=data,
    ymajorgrids=true,
    xmajorgrids=true,
    axis lines*=left,
    ]
    \addplot +[draw=ProcessBlue,fill=ProcessBlue] coordinates {(Source-domain,82.97) (Cross-domain,84.91) (Lower-resolution,73.97) (Higher-resolution,83.56)};
    \addlegendentry{nnUNet (126.6M)}
    
    \addplot +[draw=BrickRed,fill=BrickRed] coordinates {(Source-domain,82.88) (Cross-domain,74.59) (Lower-resolution,82.39) (Higher-resolution,83.19)};
    \addlegendentry{MedSAM (4.1M)}
    
    \addplot +[draw=BurntOrange, fill=BurntOrange] coordinates {(Source-domain,85.05) (Cross-domain,70.10) (Lower-resolution,81.26) (Higher-resolution,84.33)};
    \addlegendentry{SwIPE (2.7M)}
    
    \addplot +[draw=ForestGreen, fill=ForestGreen] coordinates {(Source-domain,91.65) (Cross-domain,88.83) (Lower-resolution,91.45) (Higher-resolution,91.33)};
    \addlegendentry{I-MedSAM (1.6M)}
    \end{axis}
    \end{tikzpicture}
    
    \caption{Segmentation quality comparison between our I-MedSAM and baselines. }
    \label{fig:short-b}
  \end{subfigure}

\caption{(a) Continuous representation with implicit decoders exhibits superior scale flexibility. (b) I-MedSAM with the fewest trainable params (1.6M) surpasses the state-of-the-art discrete and implicit approaches and exhibits a solid generalization ability when facing data shifts. Please refer to \cref{sec:exp} for more experiment details.} 
\label{fig:intro}
\end{figure}
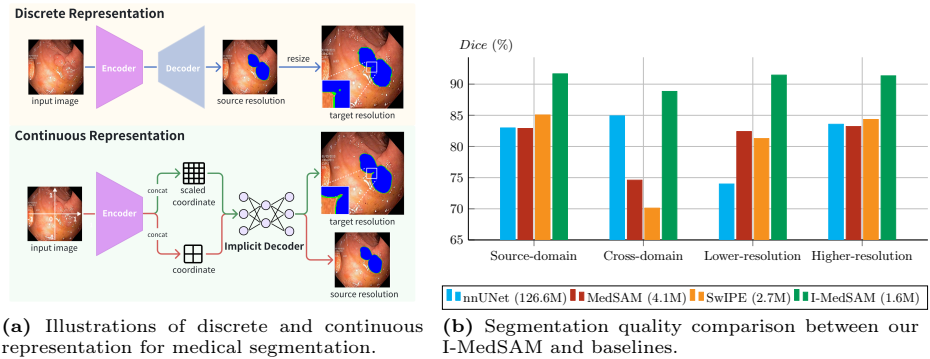

Medical image segmentation, as a pivotal component of auxiliary disease diagnosis, holds a crucial role in medical image applications. The advent of deep learning has spurred the widespread adoption of neural networks customized for medical images. For example, nnUNet~\cite{ronneberger2015unet} leverages the downsampling and upsampling modules to aggregate multi-scale contextual features. Transformers~\cite{vaswani2017attention} uses the self-attention mechanism to significantly augment the representation capacity of deep neural networks, improving the accuracy in medical image segmentation~\cite{chen2021transunet}. Recent advancements have witnessed the integration of foundation models as backbones in various works. The Segment Anything Model (SAM)~\cite{kirillov2023segment} demonstrates unprecedented zero-shot segmentation ability. Therefore, diverse adapters based on parameter-efficient fine-tuning (PEFT) are crafted to fine-tune SAM for medical images~\cite{MedSAM, samed, medical_sam_adapter, wei2023noc}.

Despite their notable effectiveness, these methods primarily focus on pixel-wise or voxel-wise predictions~\cite{ronneberger2015unet, fan2020pranet, gao2019res2net, isensee2021nnu, MedSAM, samed}. While they achieve promising results, the discrete representations present challenges in spatial flexibility and introduce discretization artifacts when scaling to arbitrary input sizes. Additionally, the discrete representations give rise to ambiguity when extracting the nuanced details crucial for precise boundary delineation~\cite{mildenhall2021nerf}, which is important in medical image analysis. The delineation of boundaries can signify the transitions between different human tissues or anatomical structures, thus providing essential information for accurately separating these instances. Usually, the intricacy and subtlety of this delineation process need additional refinement~\cite{pasupathy2015neural, alahmadi2023boundary}.


Compared with discrete representations, continuous representations learn Implicit Neural Representations (INRs)~\cite{molaei2023implicit} to transform discrete representations into continuous space. As shown in Fig.~\ref{fig:short-a}, numerous approaches learn a mapping from encoded image features and grid coordinates to the segmentation output, enabling adaptability to various output resolutions~\cite{zhang2023swipe, Khan2022IOSNet, Reich2021OSSNetME, naval2021implicit}. However, current approaches show unsatisfying domain transfer ability due to the limited representation capabilities of their pre-trained image encoders. Additionally, the boundary information of images demonstrates a strong correlation with features in the frequency domain~\cite{feng2023diffdp, li2023zero}, which is also ignored by most previous methodologies. Lastly, existing methods adopt random sampling across coordinates, underestimating the influence of sampling strategies when learning INRs. 

To address the aforementioned limitations, we propose I-MedSAM, a model that leverages the benefits of both continuous representations and SAM, aiming to enhance cross-domain capabilities and achieve precise boundary delineation. Given the medical images, I-MedSAM extracts the features from SAM with the proposed frequency adapter, which aggregates high-frequency information from the frequency domain. These features along with the grid coordinates are decoded into segmentation outputs by the learned INRs. We employ a two-stage implicit segmentation decoder, consisting of two INRs in a coarse-to-fine manner. The first INR produces coarse segmentation results and features. Subsequently, a novel uncertainty-guided sampling strategy is applied to sample \textit{Top-K variance} feature points along with their corresponding grid coordinates. Finally, these selected samples are fed into the second INR to obtain refined segmentation results. Notably, I-MedSAM is trained end-to-end with a minimal number of trainable parameters, yet it achieves state-of-the-art performance compared with all baseline methods. Our main contributions are summarized as follows:



\begin{itemize}
\item We propose I-MedSAM, a novel method that leverages the advantages of SAM and continuous representations. 

\item We propose a novel frequency adapter that utilizes high-frequency information to enhance features, thereby accurately segmenting boundaries.

\item We propose a novel coarse-to-fine INR decoder with an uncertainty-guided sampling (UGS) strategy, to learn a mapping from features and coordinates to segmentation output. 

\item We perform detailed evaluations of I-MedSAM on 2D medical image segmentation. As shown in Fig.~\ref {fig:short-b}, I-MedSAM outperforms state-of-the-art continuous and discrete methods. Experiments also demonstrate that I-MedSAM is robust to scale and domain shifts.
\end{itemize}

\section{Related Work}

\textbf{Implicit Neural Representation.} The concept of signal representation is fundamental across various domains, especially in the field of computer vision~\cite{mildenhall2021nerf, molaei2023implicit}. Traditional methods for encoding signals discretize the input space into pixel or voxel grids~\cite{ronneberger2015unet, fan2020pranet, hatamizadeh2022unetr,isensee2021nnu, samed, cheng2024unleashing, cheng2023sam, bui2023sam3d, deng2023sam, shi2024mask}. Different from these discrete methods, Implicit Neural Representation (INR) learns generator functions that map input coordinates into the signal values~\cite{chibane2020implicit}. Numerous studies employ INR for diverse tasks, including medical data reconstruction, rendering, compression, registration, super-resolution and segmentation~\cite{molaei2023implicit, yang2022implicitatlas, amiranashvili2022learning, mcginnis2023single, kuang2022makes}. For segmentation, conventional methods typically consist of a trained feature encoder and a decoder. The encoder encodes medical data into features, and the decoder subsequently decodes features along with their coordinates into segmentation output~\cite{khan2022implicit, Reich2021OSSNetME, hu2022ifanet, Khan2022IOSNet, Srensen2022NUDF, zhang2023swipe, you2023implicit, stolt2023nisf}. However, current methods exhibit an imbalance in emphasizing either global or local features and demonstrate relatively low out-of-distribution ability. In contrast, our proposed approach leverages the segmentation foundation model SAM to enrich feature extraction. Moreover, we introduce an innovative uncertainty-guided sampling (UGS) strategy into the INR, enabling the adaptive selection of samples to train the implicit segmentation decoder. Please refer to the appendix for more related work.

\section{Methodology}

In this section, we initially provide a concise overview of the implicit image segmentation problem. Then, we proceed to elaborate on the pipeline of I-MedSAM. Finally, we elucidate the novel designs introduced in I-MedSAM.

\begin{figure*}[!htbp]
    \centering
    \includegraphics[width=0.70\textwidth]{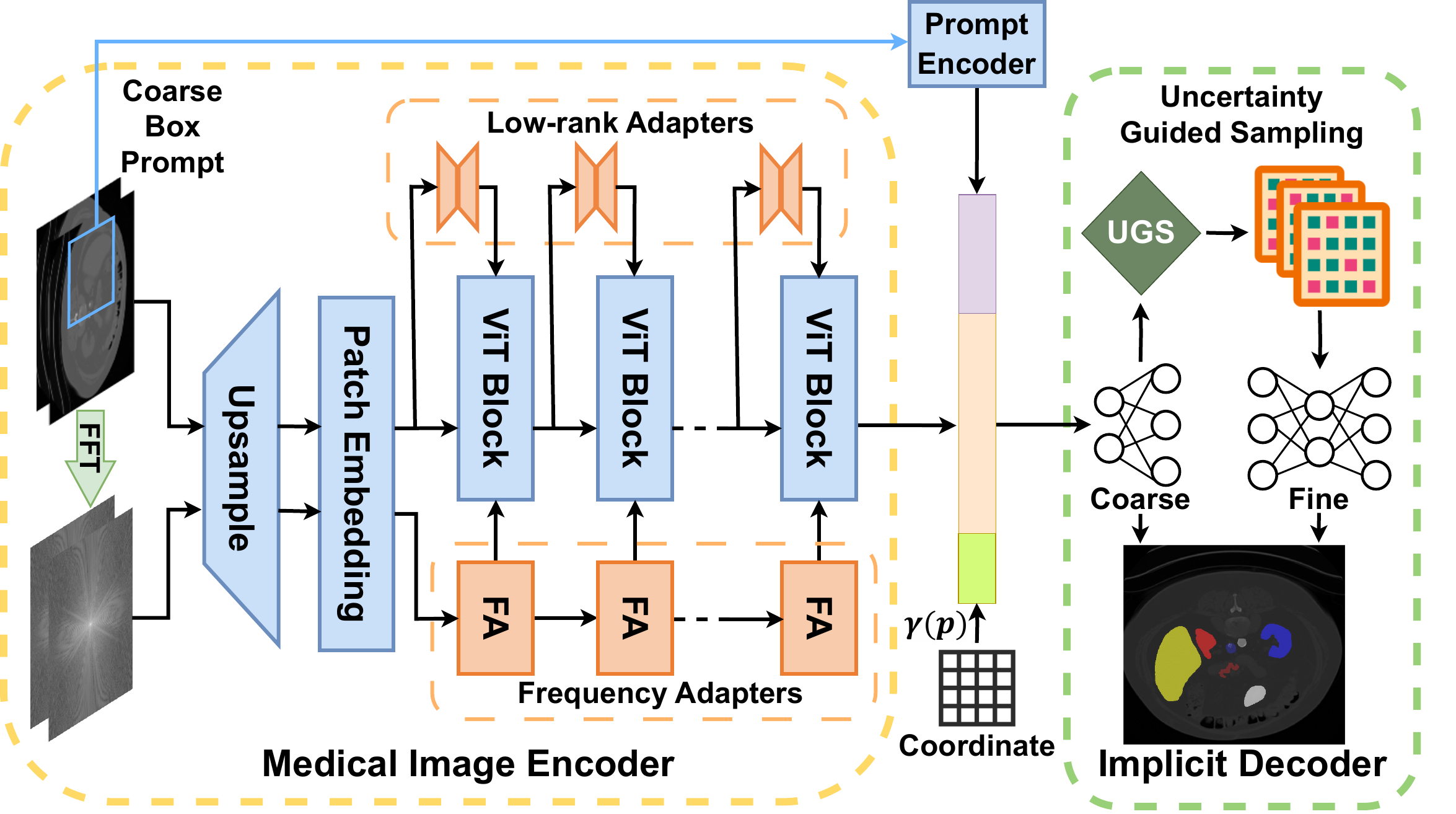}
    \caption{The overall pipeline of I-MedSAM. First, given the medical images and a coarse bounding box as a prompt, I-MedSAM utilizes the medical image encoder and the prompt encoder to generate discrete features. For the medical image encoder, we design low-rank adapters and frequency adapters to extract information from the spatial domain and frequency domain. Then I-MedSAM interpolates all features to align with the encoded coordinates and decodes them in coarse to fine neural fields. We propose an Uncertainty Guided Sampling (UGS) strategy to adaptively choose the highest variance points and refine predictions. I-MedSAM merges the predictions from coarse and fine neural fields as the final prediction maps.}
    \label{fig:pipeline}
\end{figure*}

\subsection{Preliminaries}
In traditional discrete segmentation with $C$ classes, neural networks aim to learn a direct mapping from input medical images $X \in \mathbb{R}^{H \times W \times 3}$ to class probabilities $O \in \mathbb{R}^{H \times W \times C}$ at the same resolution, in which $H$ and $W$ means height and width of input images, respectively.

On the other hand, implicit image segmentation seeks to map each pixel of medical images $X$ with its coordinate $p_{i} = (x, y)$, where $x, y \in [-1,1]$, to class probabilities $\hat{o}_{i} \in \mathbb{R}^C$, denoted as $\mathcal{N}_\theta: (p_{i}, X_{i}) \rightarrow \hat{o}_{i}$. Here, $\mathcal{N}_\theta$ represents a neural network parameterized by weights $\theta$. This formulation incorporates coordinates directly on pixels, adjusting the spatial granularity of input coordinates for predictions at arbitrary resolution, from source resolution $H\times W$ to target resolution $H'\times W'$, which can be represented as $X \in \mathbb{R}^{H \times W \times 3} \rightarrow O \in \mathbb{R}^{H' \times W' \times C}$. Moreover, it allows the direct application of pixel-wise loss functions like Cross Entropy or Dice. Additionally, the zero-isosurface in $\mathcal{N}_\theta$'s implicit space represents object boundaries, providing an additional advantage for boundary modeling.

\subsection{Overall Pipeline}
As depicted in \cref{fig:pipeline}, I-MedSAM comprises two main parts. The first part integrates an image encoder with its adapters, forming $Enc_I$, and a prompt encoder $Enc_P$, following SAM's design. Specifically, recognizing the significant role of the frequency domain in segmentation boundary representation, a frequency adapter is devised for extracting frequency features. Taking a medical image and a prompt bounding box as inputs, multi-scale features are extracted from both spatial and frequency domains. 
In scenarios involving cross-resolution, the extracted features need to be interpolated from the source resolution to achieve segmentation output at the target resolution.

The second part is the implicit segmentation decoder $Dec$, comprising two stacked INRs: one "coarse" $Dec_c$ with shallow layers and one "fine" $Dec_f$ with deeper layers. Typically, $Dec_c$ generates a coarse segmentation map, and $Dec_f$ refines it on sampled points. The selection of these points is determined by the pixel-wise uncertainty of segmentation predictions, assessed through MC-Dropout and Top-K algorithms. Detailed explanations of these two parts will be provided in the following sections.


\subsection{Medical Image Encoder}
In this section, we introduce the frequency adapter and low-rank adapter integrated into SAM, to extract features from both frequency and spatial domains. 

\textbf{Frequency Adapter.} 
Discrete Fourier Transform (DFT) is a common and effective method for transforming an image into the frequency domain. In practice, the Fast Fourier Transform (FFT) is employed for efficient computation of DFT, the spectrum representation of $f_{h,w}$ can be formulated as:
\begin{equation}
    \mathcal{F}_{u,v} = \sum_{h=1}^{H} \sum_{w=1}^{W} f_{h,w} \cdot \mathrm{e}^{-j2\pi \left(\frac{h}{H}u+\frac{w}{W}v\right)}.
\end{equation}
Subsequently, the amplitude and phase spectrum of $\mathcal{F}_{u, v}$ can be obtained as $|\mathcal{F}_{u,v}|$ and $arg(\mathcal{F}_{u,v})$, respectively. Experiment results in Tab. \ref{tab:ablation_freq_ada} indicate that the amplitude spectrum exhibits superior representation ability compared to the phase spectrum. Therefore, we default to using the amplitude spectrum for our proposed frequency adapter (FA).

\begin{wrapfigure}{r}{0.45\textwidth}
    \centering
    \includegraphics[width=0.4\textwidth]{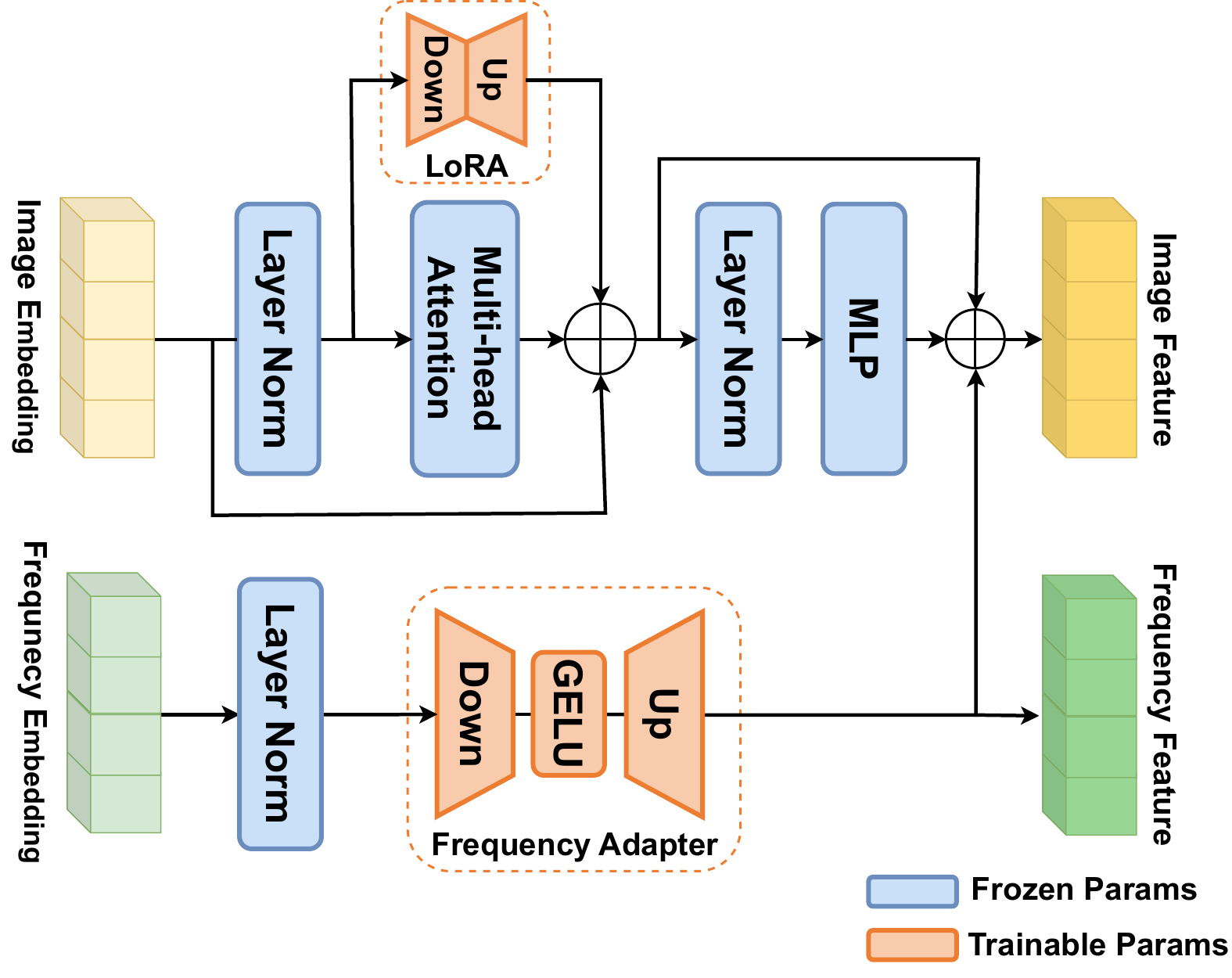}
    \caption {Illustration of the proposed frequency adapter and LoRA in the image encoder. The image/frequency embedding from patch embedding undergoes two separate branches in the encoder. 
    }
    \label{fig:frequency_adapter}
\end{wrapfigure}

As illustrated in \cref{fig:frequency_adapter}, the individual FA comprises a linear down-projection layer, a GELU activation layer, and a linear up-projection layer. In total, we utilize $n$ instances of FA as a sequence, corresponding to the number of Vision Transformer (ViT) Blocks of $Enc_I$. 

\textbf{Low-Rank Adapter.} In contrast to fine-tuning all parameters in the image encoder $Enc_I$, we leverage the Low-Rank Adapter (LoRA)~\cite{hu2021lora} to update a small fraction of parameters, adapting SAM to medical images, as illustrated in \cref{fig:frequency_adapter}. 
Given the encoded token sequence $F\in \mathbb{R}^{B\times N\times C_{in}}$, the resulting token sequence $\hat{F}\in \mathbb{R}^{B\times N\times C_{out}}$ is generated using a projection layer $W_{p}\in \mathbb{R}^{C_{out}\times C_{in}}$, denoted as $\hat{F} = W_{p}F$. LoRA proposes that the adjustment to $W_p$ should be gradual and consistent. It recommends utilizing a low-rank approximation $A\in \mathbb{R}^{r\times C_{in}}$ and $B\in \mathbb{R}^{C_{out} \times r}$ to represent this gradual update, which can be formulated as:
\begin{equation}
\hat{W_{p}} = W_{p}+\Delta W_{p}=W_{p}+BA.
\end{equation}

As the multi-head attention mechanism determines the regions to focus on, it is reasonable to apply LoRA to the frozen projection layers of query $Q$, key $K$, or value $V$ to influence the attention scores. We notice that I-MedSAM performs better when LoRA is applied to the query $Q$ and value $V$ projection layers, which can be expressed as:
\begin{equation}
    \left\{
        \begin{aligned}
            Q &= \hat{W_q}F=W_qF+B_qA_qF \\
            K &= W_kF \\
            V &= \hat{W_v}F=W_vF+B_vA_vF   
        \end{aligned}
    \right.
\end{equation}
where $W_q$, $W_k$ and $W_v$ are frozen projection layers from SAM's image encoder, and $A_q$, $B_q$, $A_v$, $B_v$ are trainable LoRA parameters.



\subsection{Implicit Segmentation Decoder}
In this section, we introduce a coarse-to-fine implicit neural representation with an uncertainty-guided sampling (UGS) strategy to decode features from encoders into the segmentation maps at target resolutions.

\textbf{Coarse to Fine Implicit Neural Representation.} 
Given features from the image encoder $Enc_I$ and the prompt encoder $Enc_P$, we interpolate them from source to target resolutions and concatenate them with coordinates $p$. These coordinates $p$ are generated at the target resolutions and normalized to $[-1,1]$. To address potential biased learning resulting from the direct use of input coordinates~\cite{rahaman2019spectral}, we encode the coordinates into a higher-dimensional space using a high-frequency positional encoding function, which is defined as:
\begin{equation}
        \gamma(p) = ( \sin(2^0\pi p), \cos(2^0\pi p), \cdots, 
                      \sin(2^{L-1}\pi p), \cos(2^{L-1}\pi p) \,)
\end{equation}
where the hyperparameter $L$ is set to 10 in our experiments following the previous work.
The encoded coordinates, the encoded features from both the image and prompt encoders are concatenated to feed into the decoder:
\begin{equation}
    \begin{aligned}
        Z^{p} = Concat(\gamma(p), Interp(Enc_I(X)), Enc_I(P)).
    \end{aligned}
    \label{eq:enc}
\end{equation}
Here, $X$ and $P$ represent the input medical image and the corresponding coarse bounding box prompt, respectively. The function $Interp$ refers to the interpolation function based on bilinear algorithms, which is used to interpolate 
the encoded features from source to target resolution, in alignment with the encoded coordinates.

Inspired by NeRF~\cite{mildenhall2021nerf}, we depart from the one-stage INR approaches to introduce a two-stage decoding process. This involves optimizing two INRs simultaneously: one "coarse" $Dec_c$, with shallow layers, and one "fine" $Dec_f$, with deeper layers. $Dec_c$ produces a coarse segmentation map, $\hat{o}^c_i$, serving as reference for $Dec_f$ to refine. Additionally, $Dec_c$ generates coarse features, $z^{c}_i$, employed by $Dec_f$ in its refinement process.


We employ MC-dropout to calculate the uncertainty of features $\hat{o}^c_i$ for each pixel. Subsequently, a Top-K percentage of feature points is sampled based on this uncertainty, denoted as $z^s_i$ (with $s \in \mathcal{S}$). Finally, the predictions from the "coarse" and "fine" INRs are combined to produce the output of I-MedSAM. The decoding process is formulated as follows:
\begin{equation}
    \begin{aligned}
        \hat{o}^c_i, \, z^{c}_i & = Dec_c(z^{p}_i) \\
        z^{s}_{i} & = UGS(z^{c}_i), \, s \in \mathcal{S} \\
        \hat{o}^f_i & = Dec_f(z^{s}_{i}) \\
        \hat{O} & = \hat{O}^c(\mathcal{S} \setminus s) \cup \hat{O}^f(s) \, , \hat{o}_i \in \hat{O}.
    \end{aligned}
\end{equation}
Here, $UGS$ represents Uncertainty Guided Sampling, which will be further illustrated in the following section.

\textbf{Uncertainty Guided Sampling.} 
In the sampling process, we select feature points that require refinement from the "coarse" INR $Dec_c$ and feed them into the "fine" INR $Dec_f$, based on uncertainty estimation. Drawing inspiration from MC-Dropout methods~\cite{YarinGal2015DropoutAA, ChuanGuo2017OnCO}, we apply dropout $T$ times to obtain $T$ prediction results of coarse segmentation probabilities, $\{o_i^c\}_{t=1}^T$, given the input features $z_i^p$, denoted as $\{p_t(o^c_i|z^{p}_i)\}_{t=1}^T$. The uncertainty is calculated as the variance of predictions for each feature point, expressed as:
\begin{equation}
\label{Var_uncertainty}
\left\{
    \begin{aligned}
        \mu_i &= \frac{1}{T}\sum_{t=1}^T(p_t(o^c_i|z^{p}_i)) \\
        u_i &= \frac{1}{T}\sum_{t=1}^T({p_t(o^c_i|z^{p}_i)}-\mu_i)^2.
    \end{aligned}
\right.
\end{equation}
Subsequently, we sample the feature points with the highest Top-K percentage uncertainty to form $z_i^s$ for $Dec_f$ to refine. This estimation of uncertainty reflects the variation in prediction difficulty among different samples. It adaptively selects pixels with higher difficulty for refinement by $Dec_f$, achieving more accurate segmentation results. 

\subsection{Training I-MedSAM}

To optimize the trainable parameters of I-MedSAM, we freeze the pre-trained image encoder $Enc_I$, while only unfreezing the proposed adapters, prompt encoder $Enc_P$ and INRs. 
We utilize SAM's image encoder with LoRA and our proposed frequency adapter to extract features for input medical images $X$, while extracting prompt features in terms of coarse bounding box $P$ for targeted segmentation objects. The coarse bounding box $P$ is randomly adjusted the height and width following the previous work~\cite{samed}.
Then we concatenate features along with the mapped coordinate values and decode them with the proposed two-stage INR decoder. With the coarse to fine INR and the uncertainty guided sampling strategy, I-MedSAM obtains the coarse and refined point-wise segmentation probabilities $\{\hat{o}^{c}_i, \hat{o}^{f}_i\}_{i \in X}$, combined as $\hat{o_i}$. For training optimization, we adopt pixel-wise segmentation loss, which can be formulated as: 
\begin{equation}
\begin{aligned}
L_{seg}(o_i, \hat{o}_i) = 0.5 \cdot L_{ce}(o_i, \hat{o}_i) + 0.5 \cdot L_{dc}(o_i, \hat{o}_i)
\end{aligned}
\end{equation}
where $L_{ce}$ and $L_{dc}$ stand for Cross Entropy loss and Dice loss  respectively. We apply the loss to supervise both coarse and refined segmentation maps progressively. Within the training process, we decrease weights for coarse supervision and increase weights for refined supervision until I-MedSAM converges.

\section{Experiments}
\label{sec:exp}
In this section, we present extensive experiments to evaluate the effectiveness of I-MedSAM for medical image segmentation. We first introduce the experimental settings including datasets and training details. Then we compare our method with the SOTA implicit and discrete approaches on the binary polyp segmentation~\cite{jha2021sessilecomprehensive} and multi-class organ segmentation~\cite{bcv2015} qualitatively and quantitatively. We further evaluate the performance and robustness of I-MedSAM when facing data shifts. Finally, we conduct a comprehensive ablation study to evaluate the contribution of each component. Due to space limitations, we provide more details and visualization results in the supplementary material.


\begin{table}[t]
    \begin{center}
    \caption{
        \label{tab:overall}
        Overall segmentation results versus the state-of-the-art discrete approaches and implicit approaches. The Trainable Params columns report unfrozen parameters in training and the Dice columns report averaged scores with standard deviation.
    }
    \resizebox{0.90\linewidth}{!}{
    \begin{tblr}{
        columns={colsep=7pt},
        colspec={l c c | l c c},
    }
    \hline 
    \SetCell[c=3]{c} Binary Polyp Segmentation & & & \SetCell[c=3]{c} Multi-class Organ Segmentation \\
    \cline{1-3} \cline{4-6}
    Method & Dice (\%)$\uparrow$ & Trainable Params (M)$\downarrow$ & Method & Dice (\%)$\uparrow$ & Trainable Params (M)$\downarrow$ \\
    
    \hline 
    \SetCell[c=6]{c} \textit{Discrete Approaches} \\
    \hline 
    
    \makebox[\nameblob][l]{U-Net}
    \makebox[\blob][r]{\cite{ronneberger2015unet}} & 63.89±1.30 & 7.9  & 
    \makebox[\nameblob][l]{U-Net}
    \makebox[\blob][r]{\cite{ronneberger2015unet}} & 74.47±1.57 & 16.3 
    \\
    
    \makebox[\nameblob][l]{PraNet} \makebox[\blob][r]{\cite{fan2020pranet}} & 82.56±1.08 & 30.5  &
    \makebox[\nameblob][l]{UNETR} \makebox[\blob][r]{\cite{hatamizadeh2022unetr}} & 81.14±0.85 & 92.6 
    \\
    
    \makebox[\nameblob][l]{Res2UNet} \makebox[\blob][r]{\cite{gao2019res2net}} & 81.62±0.97 & 25.4  &
    \makebox[\nameblob][l]{Res2UNet} \makebox[\blob][r]{\cite{gao2019res2net}} & 79.23±0.66 & 38.3 
    \\
    
    \makebox[\nameblob][l]{nnUNet} \makebox[\blob][r]{\cite{isensee2021nnu}} & 82.97±0.89 & 126.6  &
    \makebox[\nameblob][l]{nnUNet} \makebox[\blob][r]{\cite{isensee2021nnu}} & 85.15±0.67 & 126.6  \\

    \makebox[\nameblob][l]{MedSAM} \makebox[\blob][r]{\cite{MedSAM}} & 82.88±0.55 & 4.1  &
    \makebox[\nameblob][l]{MedSAM} \makebox[\blob][r]{\cite{MedSAM}} & 85.85±0.81 & 52.7  \\
    \hline 
    \SetCell[c=6]{c} \textit{Implicit Approaches}\\
    \hline

    \makebox[\nameblob][l]{OSSNet} \makebox[\blob][r]{\cite{Reich2021OSSNetME}} & 76.11±1.14 & 5.2  & 
    \makebox[\nameblob][l]{OSSNet} \makebox[\blob][r]{\cite{Reich2021OSSNetME}} & 73.38±1.65 & 7.6  \\

    \makebox[\nameblob][l]{IOSNet} \makebox[\blob][r]{\cite{Khan2022IOSNet}} & 78.37±0.76 & 4.1  & 
    \makebox[\nameblob][l]{IOSNet} \makebox[\blob][r]{\cite{Khan2022IOSNet}} & 76.75±1.37 & 6.2  \\

    \makebox[\nameblob][l]{SwIPE} \makebox[\blob][r]{\cite{zhang2023swipe}} & 85.05±0.82 & 2.7  &
    \makebox[\nameblob][l]{SwIPE} \makebox[\blob][r]{\cite{zhang2023swipe}} & 81.21±0.94 & 4.4  \\

    I-MedSAM (ours) & \textbf{91.49}±0.52 & \textbf{1.6}  &
    I-MedSAM (ours) & \textbf{89.91}±0.68 & \textbf{3.5}  \\

    \hline 
    \end{tblr}
    }
    \end{center}
\end{table}

\begin{figure*}[t]
    \centering
    \includegraphics[width=0.7\textwidth]{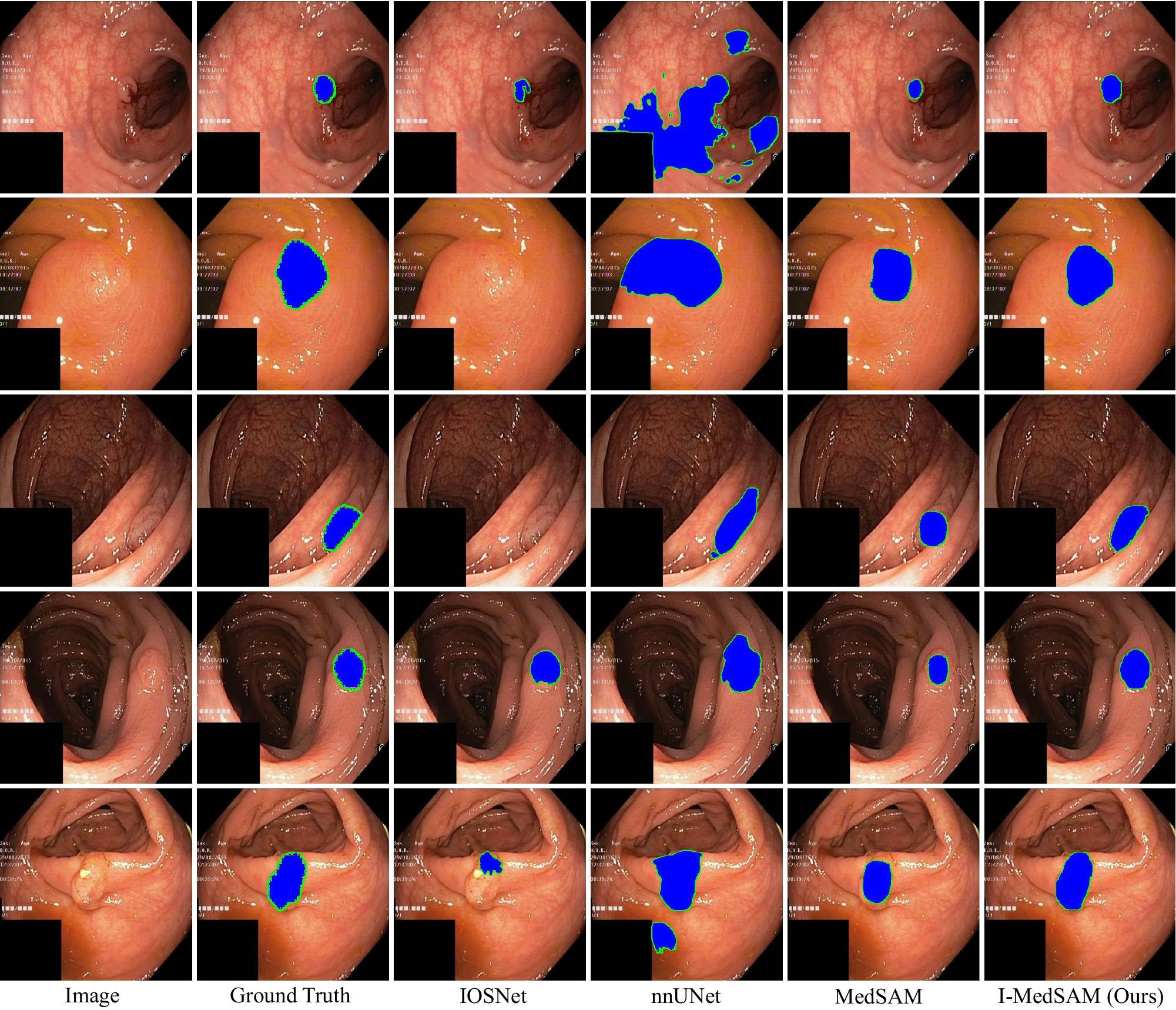}
    \caption{Qualitative comparison on Kvasir-Sessile dataset for binary polyp segmentation. }
    \label{fig:Kvasir_visual}
\end{figure*}

\begin{figure*}[thbp]
    \centering
    \includegraphics[width=0.7\textwidth]{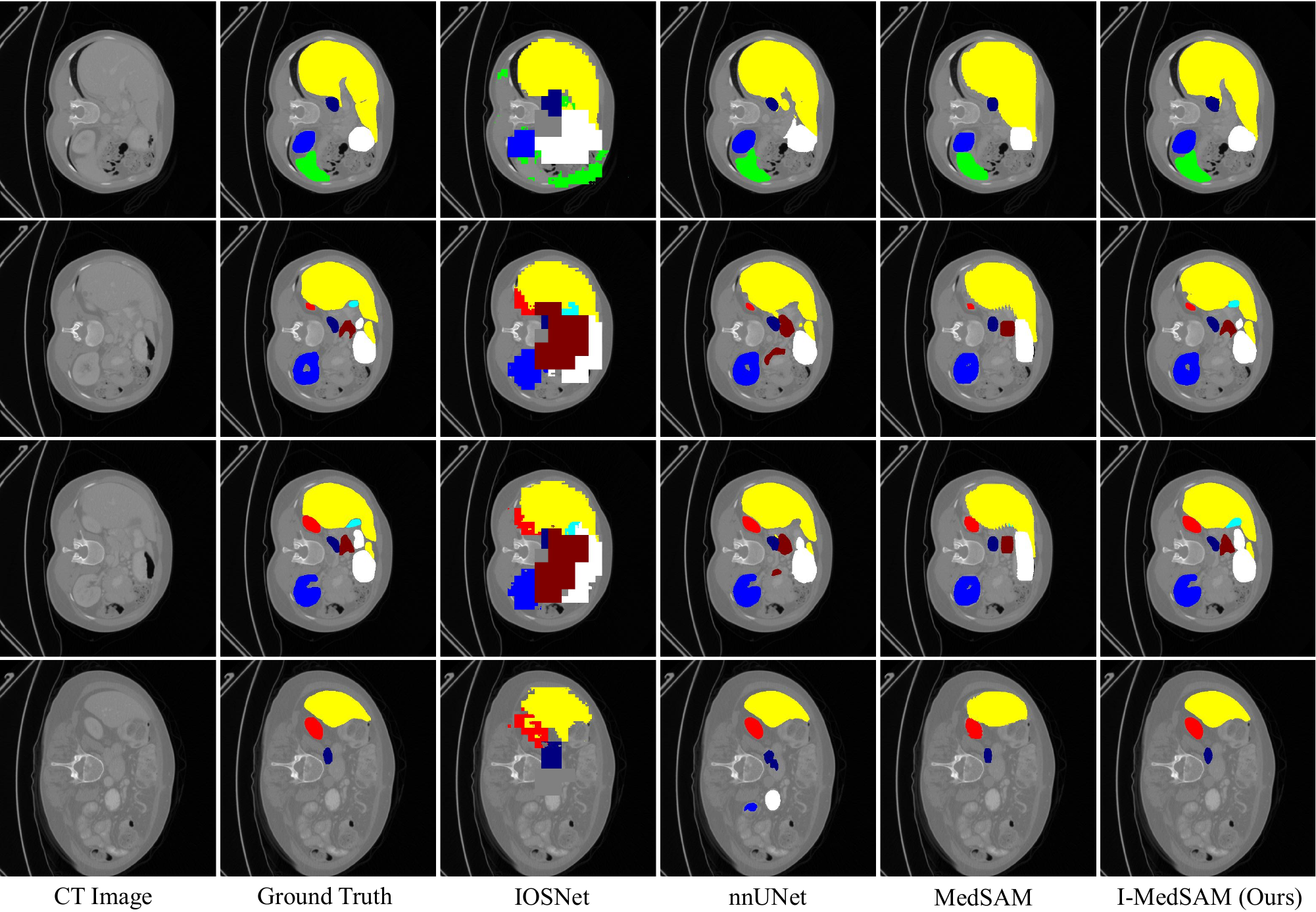}

    \caption{Qualitative comparison on BCV dataset for multi-organ segmentation. }
    \label{fig:BCV_visual}
\end{figure*}

\subsection{Experimental Settings}

\textbf{Datasets.} We assess the performance of our model on two distinct tasks: binary polyp segmentation and multi-class abdominal organ segmentation. For binary polyp segmentation, we conduct experiments using the challenging \textbf{Kvasir-Sessile} dataset \cite{jha2021sessilecomprehensive}, consisting of 196 RGB images of small sessile polyps. Additionally, we evaluate the generalization capability of our model by testing the pre-trained I-MedSAM directly to the \textbf{CVC-ClinicDB} dataset \cite{bernal2015cvc}, comprising 612 images from 31 colonoscopy sequences.

For multi-organ segmentation, training is conducted on the \textbf{BCV} dataset \cite{bcv2015}, comprising 30 CT scans with annotations for 13 organs. Model robustness is also evaluated using the pre-trained I-MedSAM on diverse CT images in \textbf{AMOS} \cite{ji2022amos} (200 training CTs, maintaining the same setup as in~\cite{zhang2022spade}). Since our work is dedicated to showing the effectiveness of 2D medical image segmentation, we just slice-wise segment CT data. Following the data prepossess in SwIPE\cite{zhang2023swipe}, all datasets are divided with a train:validation:test ratio of 60:20:20, and the reported dice scores are all based on the test set.

\textbf{Implementation Details.} The training process of I-MedSAM involves fine-tuning the encoder of SAM~\cite{kirillov2023segment}, utilizing ViT-B as the backbone. We set LoRA ranks to $4$ and incorporate amplitude information in the frequency adapters. 
For implicit segmentation decoders, we set latent MLP dimensions as $[1024, 512]$ for $Dec_{c}$ and $[512, 256, 256, 128]$ for $Dec_{f}$. We sample the highest uncertainty points with a proportion of $12.5\%$ and set the dropout probabilities to $0.5$. 

For multi-organ segmentation, we slightly modify the number of the last layer in $Dec_{c}$ and $Dec_{f}$ to match target segmentation classes. I-MedSAM are optimized by AdamW~\cite{Loshchilov2017DecoupledWDAdamW} with $\alpha$=$0.5$, $\beta$=$0.1$, and $\lambda_{ada}$=$5\times 10^{-5}$ for adapters in the encoder and $\lambda_{dec}$=$1\times 10^{-3}$. For fair comparisons, all methods are trained for $1000$ epochs on the same experiment settings. The reported test dice score and Hausdorff Distance~\cite{huttenlocher1993comparing} correspond to the best validation epoch. Image input sizes are $384\times384$ for Sessile and $512\times512$ for each slice of BCV.

\textbf{Baselines.} We divide baselines into two sets: discrete approaches and implicit (continuous) approaches. 
Discrete approaches include U-Net~\cite{ronneberger2015unet}, PraNet~\cite{fan2020pranet}, Res2UNet~\cite{gao2019res2net}, nnUNet~\cite{isensee2021nnu}, UNETR~\cite{hatamizadeh2022unetr} and MedSAM~\cite{MedSAM}. Specifically, MedSAM~\cite{MedSAM} is also a SAM-based approach, where SAM's original decoder is directly finetined. 
Implicit approaches include OSSNet~\cite{Reich2021OSSNetME}, IOSNet~\cite{Khan2022IOSNet} and SwIPE~\cite{zhang2023swipe}. Given the absence of code availability from the prior state-of-the-art SwIPE~\cite{zhang2023swipe} for extended comparison and the demonstrated superiority of IOSNet~\cite{Khan2022IOSNet} over OSSNet~\cite{Reich2021OSSNetME} in implicit approaches, we primarily compare our method with IOSNet~\cite{Khan2022IOSNet} in several experiment settings.

\subsection{Quantitative Comparison}
In this section, we first report the dice score compared with baselines. Then we conduct experiments across different resolutions and domains to evaluate the robustness and generalization ability under data shifts. Finally, we implement Hausdorff Distance (HD distance)~\cite{huttenlocher1993comparing} to compare the quality of the segmentation boundary with baselines across different experiment settings.



\textbf{Segmentation Comparison.} We make a comparison with both the discrete methods and implicit methods in terms of trainable parameters and Dice score with standard deviation on Polyp Sessile for binary-class segmentation and CT BCV for multi-class segmentation, as demonstrated in Tab.~\ref{tab:overall}. On the smaller polyp dataset, we observe notable improvements over the best-known implicit methods and discrete methods with much fewer trainable parameters (1.26\% of nnNet~\cite{isensee2021nnu} and 59.26\% of SwIPE~\cite{zhang2023swipe}). For multi-organ segmentation on BCV, performance gains are also significant compared with the best-known implicit methods and discrete methods. On the one hand, SAM, with the assistance of the proposed frequency adapters, generates abundant features that enhance the quality of segmentation boundaries. In contrast, SwIPE employs Res2Net-50~\cite{gao2019res2net} as its backbone, which offers less sufficient features compared to SAM's, leading to inferior segmentation quality. On the other hand, with the proposed uncertainty-guided sampling in INR decoders, I-MedSAM adaptively selects and refines uncertain pixels with the highest variance, leading to more accurate segmentation maps.

\begin{wraptable}{r}{0.4\textwidth}
    \centering
    \caption{ 
        \label{tab:across_resol}
        Cross-resolution from $384\times 384$ to $128\times 128$, from $384\times 384$ to $896 \times 896$ on Kvasir-Sessile. 
    }
    \resizebox{0.37\textwidth}{!}{
    \begin{tblr}{
        colspec={l | c c}
    }
    \hline
    \SetCell[r=2,c=1]{c} Method & \SetCell[c=2]{c} Dice (\%)$\uparrow$ \\
    \cline{2-3}
    & 384 $\rightarrow$ 128 & 384 $\rightarrow$ 896 \\
    \hline  
    \SetCell[c=3]{c} \textit{Discrete Approaches} \\
    \hline
    \makebox[\nameblob][l]{PraNet} \cite{fan2020pranet} & 72.64 & 74.95 \\
    \makebox[\nameblob][l]{PraNet*} \cite{fan2020pranet} & 68.79 & 43.92 \\
    \makebox[\nameblob][l]{nnUNet} \cite{isensee2021nnu} & 73.97 & 83.56 \\
    \makebox[\nameblob][l]{nnUNet*} \cite{isensee2021nnu} & 65.34 & 76.36 \\
    \makebox[\nameblob][l]{MedSAM} \cite{MedSAM} & 82.39 & 83.19 \\
    \makebox[\nameblob][l]{MedSAM*} \cite{MedSAM} & 82.37 & 83.32 \\
    \hline
    \SetCell[c=3]{c} \textit{Implicit Approaches} \\
    \hline
    \makebox[\nameblob][l]{IOSNet} \cite{Khan2022IOSNet} & 76.18 & 78.01 \\
    \makebox[\nameblob][l]{SwIPE} \cite{zhang2023swipe} & 81.26 & 84.33 \\
    I-MedSAM (ours) & \textbf{91.45} & \textbf{91.33} \\
    \hline
    \end{tblr}
    }
    \hfill
\end{wraptable}



\textbf{Robustness under Data Shifts.} We compare the robustness across resolutions and domains with the best discrete and implicit methods on binary-class polyp segmentation. 

Firstly, to adapt the pre-trained I-MedSAM model, originally trained on standard $384\times384$ images, to different target resolutions such as $128\times128$ for lower resolutions and $896\times896$ for higher resolutions, the input coordinates ($384\times384$) are scaled accordingly to match the target resolutions. Subsequently, the dice score is computed at these corresponding target resolutions.
For discrete methods, the resolution of output images remains the same as that of the input images. We input the original images at their source resolution ($384 \times 384$) and resize the output segmentation maps to the target resolution for evaluation. Additionally, we denote discrete baselines with the suffix $*$, where the original medical images are resized to the target resolution and then provided as input to these methods to directly generate segmentation at the target resolution.


\begin{wraptable}{r}{0.3\textwidth}
    \centering
    \caption{  
        \label{tab:across_dataset}
        Cross-domain on binary polyp segmentation and multi-class abdominal organ segmentation.
    }
    \resizebox{0.24\textwidth}{!}{
    \begin{tblr}{
        colspec={l | c}
    }
    \hline
    \SetCell[r=1,c=1]{c} Method & Dice (\%) \\
    \hline
    \SetCell[c=2]{c} \textit{Kvasir-Sessile $\rightarrow$ CVC} \\
    \hline
    \makebox[\nameblob][l]{PraNet} \cite{fan2020pranet} & 68.37 \\
    \makebox[\nameblob][l]{nnUNet} \cite{isensee2021nnu} & 84.91 \\
    \makebox[\nameblob][l]{MedSAM} \cite{MedSAM} & 74.59 \\
    \makebox[\nameblob][l]{IOSNet} \cite{Khan2022IOSNet} & 59.42 \\
    \makebox[\nameblob][l]{SwIPE} \cite{zhang2023swipe} & 70.10 \\
    I-MedSAM (ours) & \textbf{88.83} \\
    \hline
    \SetCell[c=2]{c} \textit{BCV $\rightarrow$ AMOS} \\
    \hline
    \makebox[\nameblob][l]{UNETR} \cite{fan2020pranet} & 81.75 \\
    \makebox[\nameblob][l]{nnUNet} \cite{isensee2021nnu} & 79.63 \\
    \makebox[\nameblob][l]{MedSAM} \cite{MedSAM} & 71.98 \\
    \makebox[\nameblob][l]{IOSNet} \cite{Khan2022IOSNet} & 79.48 \\
    \makebox[\nameblob][l]{SwIPE} \cite{zhang2023swipe} & 82.81 \\
    I-MedSAM (ours) & \textbf{86.28} \\
    \hline
    \end{tblr}
    }
\end{wraptable}

As shown in \cref{tab:across_resol}, implicit methods exhibit spatial flexibility and consistently outperform discrete methods. Among implicit approaches, our I-MedSAM achieves the highest performance across various output resolutions. This superior performance can be attributed to the efficacy of the proposed frequency adapters and uncertainty-guided sampling, enhancing I-MedSAM's capability to provide accurate predictions at arbitrary resolutions.


Secondly, we investigate the robustness of model performance across different datasets for the same task. In the binary-class polyp segmentation task, all methods are pre-trained on the Kvasir-Sessile dataset and evaluated directly on the CVC dataset. Similarly, in the multi-class abdominal organ segmentation task, all methods are pre-trained on the BCV dataset and evaluated on the AMOS dataset, focusing solely on the liver class. As shown in \cref{tab:across_dataset}, leveraging the generalization ability of SAM, I-MedSAM outperforms the top discrete method, achieving dice scores of 88.83\% and 86.28\% respectively. For additional visualization comparisons regarding data shifts, please refer to the supplementary materials.

\textbf{Boundary Comparison.}
We further utilize the Hausdorff Distance (HD distance)\cite{huttenlocher1993comparing} to assess segmentation boundary quality. I-MedSAM achieves a lower HD distance, indicating superior boundary quality. For more detailed boundary visualizations, please refer to Fig.\ref{fig:Kvasir_visual} and the supplementary materials.

\begin{table}[htbp]
    \centering
    \caption{Hausdorff Distance comparison on various experiment settings.}
    \resizebox{0.90\linewidth}{!}{
     \begin{tblr}{
        colspec={l | c c c c c c},
        colsep=8pt, 
        rowsep=1pt, 
    }
    
    \hline
    \SetCell[r=1,c=1]{c} HD distance ($\downarrow$) & Kvasir-Sessile & Kvasir-Sessile $\rightarrow$ CVC & 384 $\rightarrow$ 128 & 384 $\rightarrow$ 896 & BCV & BCV $\rightarrow$ AMOS \\
    \hline
    \makebox[\nameblob][l]{nnUNet} \cite{isensee2021nnu} & 31.30            & 82.31 & 13.69         & 72.31             & 6.50 & 80.39 \\
    \makebox[\nameblob][l]{MedSAM} \cite{MedSAM}  & 21.53                   & 30.15 & 8.04          & 51.82             & 10.62 & 52.14 \\
    \makebox[\nameblob][l]{IOSNet} \cite{Khan2022IOSNet}  & 51.72           & 81.60 & 35.33         & 87.86             & 21.46 & 61.19 \\
    \makebox[\nameblob][l]{I-MedSAM(Ours)}    & \textbf{11.59}  & \textbf{19.76} & \textbf{7.91} & \textbf{32.77}    & \textbf{5.95} & \textbf{37.53} \\
    \hline
    \end{tblr}
    }
    \label{table:rebuttal_tab}
\end{table}


\subsection{Qualitative Comparison}

As shown in Fig.~\ref{fig:Kvasir_visual} and Fig.~\ref{fig:BCV_visual}, we conduct qualitative comparisons on Kvasir-Sessile and BCV datasets. Due to the unavailability of code for inference, SwIPE has been omitted from the visual comparison. We also provide input medical images along with corresponding ground truth segmentation masks. 
The segmentation boundaries are delineated by green lines in Fig.~\ref{fig:Kvasir_visual}. From the figures, it can be witnessed that I-MedSAM obtains better segmentation boundaries. Thanks to the proposed frequency adapters and uncertainty guided sampling techniques, I-MedSAM can efficiently aggregate high-frequency information from the input, which is beneficial to the accuracy of final segmentation maps. 
Due to the space limitation, please refer to supplementary materials for more qualitative results.


\subsection{Ablation Study}
We conduct ablation studies focusing on three aspects: component-wise ablations, the incorporation of frequency adapter, and point numbers for sampling. In each of our ablation experiments, other hyper-parameters remain consistent with the implementation details. 

\begin{table}[htbp]
    \centering
    \caption{Effectiveness of each component of the pipeline. We evaluate the Dice metric for both cross-domain and cross-resolution tasks.}
    \resizebox{0.65\linewidth}{!}{
     \begin{tblr}{
        colspec={ c | c | c | c | c | c | c },
        colsep=3pt, 
        rowsep=1pt, 
    }
    
    \hline    
    \SetCell[r=2,c=1]{c} LoRA & \SetCell[r=2,c=1]{c} FA & \SetCell[r=2,c=1]{c} INR & \SetCell[r=2,c=1]{c} Kvasir-Sessile & \SetCell[c=1]{c} Cross-domain & \SetCell[c=2]{c} Cross-resolution \\
    \cline{4-7} &&&& Kvasir-Sessile $\rightarrow$ CVC & 384 $\rightarrow$ 128 & 384 $\rightarrow$ 896 \\
    \hline
    \makebox \checkmark &  &  & 83.61 & 82.57 & 72.73 & 76.46 \\
    \hline
    \makebox \checkmark & \checkmark &  & 88.74 & 82.61 & 75.69 & 78.59 \\
    \hline
    \makebox \checkmark &  & \checkmark & 88.83 & 83.40 & 88.16 & 88.43 \\
    \hline
    \makebox \checkmark & \checkmark & \checkmark & \textbf{91.49} & \textbf{88.83} & \textbf{91.45} & \textbf{91.33} \\
    \hline
    \end{tblr}
    }
    \label{tab:component ablation}
\end{table}

\textbf{Component-wise ablations.} To demonstrate the effectiveness of each component, we conduct a component-wise ablation on Kvasir-Sessile~\cite{jha2021sessilecomprehensive}, cross-domain and cross-resolution tasks, as can be seen in Tab.~\ref{tab:component ablation}. We employ LoRA alone as a baseline for binary segmentation, obtaining a competitive performance, which is consistent with the baseline SAMed~\cite{samed}. 
Following the incorporation of frequency adapters and INR decoders separately, the model's performance exhibits improvements. It can be observed that the INR decoder exhibits a more pronounced advantage in cross-domain and cross-resolution tasks. Furthermore, simultaneously employing frequency adapters (FA) and INR decoders can achieve a synergistic effect where 1+1>2.




\begin{wraptable}{r}{0.36\textwidth}
    \centering
    \caption{
        \label{tab:ablation_freq_ada}
        Ablation study on Frequency Adapter (FA). FA$_{pha}$ stands for utilizing of the phase spectrum of DFT, while FA$_{amp}$ stands for the amplitude spectrum of DFT.
    }
    \resizebox{0.36\textwidth}{!}{
     \begin{tblr}{
        colspec={l | c c c}
    }
    \hline
    \SetCell[r=1,c=1]{c} Setting & w/o FA & FA$_{pha}$ & FA$_{amp}$ \\
    \hline
    \makebox[\nameblob][l]{Dice (\%)} & 88.83 & 90.60 & \textbf{91.49} \\
    \hline
    \makebox[\nameblob][l]{HD} & 15.44 & 12.67 & \textbf{11.59} \\
    \hline
    \end{tblr}
    }
    
\end{wraptable}

\textbf{Incorporating the frequency adapter}. Tab.~\ref{tab:ablation_freq_ada} indicates the effectiveness of the frequency adapter, and it can be observed that amplitude information is more helpful for spectrum representation compared to phase information. The results also demonstrate the segmentation boundary benefits from the frequency adapter.

\textbf{Points Number for Sampling.} Tab.~\ref{tab:ablation_sampling} represents ablation on points number for Uncertainty Guided Sampling (UGS). This experiment reveals that I-MedSAM generates high-quality segmentation masks with the help of the proposed uncertainty guided sampling method. Excessive sampling points do not necessarily improve the final segmentation results and may lead to increased memory consumption. Conversely, insufficient sampling points may limit the areas that require refinement. Therefore, a proportion of 12.5\% for UGS is deemed appropriate for I-MedSAM. The parameter can be adjusted according to the specific tasks.

\begin{table}[hbtp]
    \centering
    \caption{
        \label{tab:ablation_sampling}  
        Ablation study on the number of sampled feature points for Uncertainty Guided Sampling (UGS). "Top-K" denotes the selection of a specified proportion, K\%, from all feature points.
    }
    \resizebox{0.85\linewidth}{!}{
       \begin{tblr}{
        colspec={l | c c c c c c}
    }
    \hline
    \SetCell[r=1,c=1]{c}  Setting & w/o UGS & Top-50\% & Top-25\% & Top-12.5\% & Top-6.25\% & Top-3.125\% \\
    \hline
    \makebox[\nameblob][c]{Dice (\%)}       & 87.77 & 90.27 & 89.59 & \textbf{91.49} & 91.01 & 90.48 \\
    \makebox[\nameblob][c]{HD Distance}     & 16.15 & 13.88 & 14.12 & \textbf{11.59} & 12.99 & 14.53 \\
    \hline
    \end{tblr}
    }
\end{table}



\subsection{Effect of different training annotations}
\begin{wrapfigure}{r}{0.35\textwidth}
\centering
    \begin{minipage}[t]{0.35\textwidth}
    \begin{tikzpicture}[scale=0.5]
            \begin{axis}[
                ylabel=Dice (\%),
                xlabel=Proportion of Training Annotation Amount (\%),
                ymin=40, xtick={10,25,50,100}, ytick={40,50,60,70,80,90},
                axis lines*=left,
                ymajorgrids=true,
                xmajorgrids=true,
                tick align=outside,
                legend style={at={(0.8,0.55)},anchor=north,legend columns=1},
                y label style={at={(axis description cs:0.15,1.0)}, anchor=south, rotate=270},
                ]   

                \addplot[draw=ForestGreen,mark=x] 
                coordinates {
                    (10,88.81)
                    (25,90.17)
                    (50,90.51)
                    (100,91.49)
                };
                \addlegendentry{Ours}

                \addplot[draw=BlueViolet,mark=x] 
                coordinates {
                    (10,81.47)
                    (25,82.10)
                    (50,82.64)
                    (100,83.00)
                };
                \addlegendentry{MedSAM}

                \addplot[draw=Rhodamine,mark=x] 
                coordinates {
                    (10,70.24)
                    (25,75.36)
                    (50,79.98)
                    (100,82.97)
                };
                \addlegendentry{nnUNet}
                
                \addplot[draw=ProcessBlue,mark=x] 
                coordinates {
                    (10,58.07)
                    (25,72.91)
                    (50,81.02)
                    (100,85.05)
                };
                \addlegendentry{SwIPE}
                
                \addplot[draw=BrickRed,mark=x] 
                coordinates {
                    (10,51.19)
                    (25,70.90)
                    (50,79.89)
                    (100,82.56)
                };
                \addlegendentry{PraNet}
                
                \addplot[draw=BurntOrange,mark=x] 
                coordinates {
                    (10,41.95)
                    (25,54.31)
                    (50,74.16)
                    (100,78.37)
                };
                \addlegendentry{IOSNet}

            \end{axis}
\end{tikzpicture}

\end{minipage}%

\label{fig:training_proportion}
\caption{Effect on different proportions of training annotation amount.} 
\end{wrapfigure}
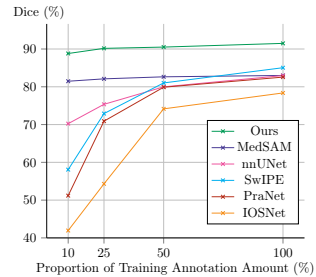

To illustrate the generalization ability between I-MedSAM and baselines, 
we further conduct experiments on different proportions of training annotation amount. 
Following the experiment settings in SwIPE~\cite{zhang2023swipe}, based on the divided training set, we train I-MedSAM on 10\%, 25\%, 50\% and 100\% of the training set. 
As shown in \cref{fig:training_proportion}, our I-MedSAM outperforms all baselines at various training annotation amounts. Thanks to the great generalization ability of SAM's encoder in I-MedSAM, I-MedSAM maintains higher segmentation performance even with relatively limited training annotations.



\section{Conclusion}



In this paper, we introduce I-MedSAM to enhance cross-domain ability and adaptability to diverse output resolutions in medical image segmentation. By integrating SAM's generalized representations into the INR space, I-MedSAM achieves state-of-the-art performance across various experimental scenarios. 
Specifically addressing the challenge of precise boundary delineation in 2D medical images, we incorporate a frequency adapter for parameter-efficient fine-tuning to SAM, showcasing the potential benefits of complementing spatial domain information with frequency domain insights for foundation models. Additionally, the employment of the uncertainty-guided sampling strategy in coarse-to-fine INRs proves effective in the selection and refinement of challenging samples in continuous space. These findings suggest avenues for future research to establish stronger connections between different representation spaces.

\section*{Acknowledgement} Shanghang Zhang is supported by the National Science and Technology Major Project of China (No. 2022ZD0117801).

%
%
\bibliographystyle{splncs04}
\bibliography{main}
\end{document}